\documentclass[sigconf]{acmart}

\AtBeginDocument{%
  }

\copyrightyear{2026}
\acmYear{2026}
\setcopyright{cc}
\setcctype{by}
\acmConference[WWW '26] {Proceedings of the ACM Web Conference 2026}{April 13--17, 2026}{Dubai, United Arab Emirates.}
\acmBooktitle{Proceedings of the ACM Web Conference 2026 (WWW '26), April 13--17, 2026, Dubai, United Arab Emirates}
\acmISBN{xxx/2026/04}
\acmDOI{10.1145/xxx.xxx}

\usepackage{amsmath}
\usepackage{amsthm}
\usepackage{hyperref}
\usepackage{url}
\usepackage{graphicx}
\usepackage{wrapfig}
\usepackage{arydshln}
\usepackage{multirow} 
\usepackage{booktabs}
\usepackage{enumerate}
\usepackage{colortbl}
\newtheorem{lemma}{Lemma}

\usepackage{algorithm}
\usepackage{algorithmic}

\begin{document}

\title{QChunker: Learning Question-Aware Text Chunking for Domain RAG via Multi-Agent Debate}

\author{Jihao Zhao}
\affiliation{%
  \institution{School of Information, Renmin University of China}
  \city{Beijing}
  \country{China}}
\email{zhaojihao@ruc.edu.cn}

\author{Daixuan	Li}
\affiliation{%
  \institution{School of Smart Governance, Renmin University of China}
  \city{Suzhou}
  \country{China}
}
\email{lidaixuan2024@ruc.edu.cn	}

\author{Pengfei	Li}
\affiliation{%
 \institution{School of Information, Renmin University of China}
 \city{Beijing}
 \country{China}}
\email{pfli9411@gmail.com}

\author{Shuaishuai Zu}
\affiliation{%
  \institution{School of Information, Renmin University of China}
  \city{Beijing}
  \country{China}}
\email{zushuaishuai@ruc.edu.cn}

\author{Biao Qin}
\authornote{Corresponding author}
\affiliation{%
  \institution{School of Information, BRAIN, Renmin University of China}
  \city{Beijing}
  \country{China}}
\email{qinbiao@ruc.edu.cn}

\author{Hongyan	Liu}
\authornotemark[1]
\affiliation{%
  \institution{School of Economics and Management, Tsinghua University}
  \city{Beijing}
  \country{China}}
\email{hyliu@tsinghua.edu.cn}

\renewcommand{\shortauthors}{Jihao Zhao, et al.}

\begin{abstract}
The effectiveness upper bound of retrieval-augmented generation (RAG) is fundamentally constrained by the semantic integrity and information granularity of text chunks in its knowledge base. Moreover, domain documents are characterized by dense terminology and strong contextual dependencies, which exacerbate the semantic fragmentation of text chunks, thereby making it difficult to efficiently utilize their key information. To address these challenges, this paper proposes \textbf{QChunker}, which restructures the RAG paradigm from retrieval-augmentation to understanding-retrieval-augmentation. Firstly, QChunker models the text chunking as a composite task of text segmentation and knowledge completion to ensure the logical coherence and integrity of text chunks. Drawing inspiration from Hal Gregersen's "Questions Are the Answer" theory, we design a multi-agent debate framework comprising four specialized components: a question outline generator, text segmenter, integrity reviewer, and knowledge completer. This framework operates on the principle that questions serve as catalysts for profound insights. Through this pipeline, we successfully construct a high-quality dataset of 45K entries and transfer this capability to small language models. Additionally, to handle long evaluation chains and low efficiency in existing chunking evaluation methods, which overly rely on downstream QA tasks, we introduce a novel direct evaluation metric, \textbf{ChunkScore}. Both theoretical and experimental validations demonstrate that ChunkScore can directly and efficiently discriminate the quality of text chunks. Furthermore, during the text segmentation phase, we utilize document outlines for multi-path sampling to generate multiple candidate chunks and select the optimal solution employing ChunkScore. Extensive experimental results across four heterogeneous domains exhibit that QChunker effectively resolves aforementioned issues by providing RAG with more logically coherent and information-rich text chunks. Notably, this study also establishes a small-domain QA dataset concerning hazardous chemical safety, which fully reveals the significant value of RAG in specialized domains and the generalization capability of the QChunker framework. Our code is available at \url{https://github.com/Robot2050/QChunker}.
\end{abstract}



\maketitle

\section{Introduction}
Retrieval-Augmented Generation (RAG) is a prominent paradigm designed to mitigate inherent limitations of Large Language Models (LLMs), such as data staleness \cite{he2022rethinking}, factual hallucinations \cite{benedict2023gen,chen2023hallucination,zuccon2023chatgpt}, and a lack of domain-specific knowledge \cite{li2023chatgpt,shen2023chatgpt}. By coupling an external retriever with a generator, RAG enhances performance on knowledge-intensive tasks, such as open-domain question answering (QA) \cite{lazaridou2022internet}, enabling more precise and allowing it to produce more accurate and context-based responses \cite{singh2021end,lin2023li}. Although the efficacy of this retrieval-based strategy is well-established, its overall performance is critically dependent on the relevance and accuracy of the retrieved documents \cite{li2022survey,tan2022tegtok}. The introduction of superfluous, irrelevant, or incomplete information through the retrieval process can fail to augment, and may even degrade, the quality of the final generated answer \cite{shi2023large,yan2024corrective}.

In fact, the relevance and accuracy of retrieved documents largely depend on the text chunking step at the source of knowledge base construction \cite{lewis2020retrieval,bhat2025rethinking,wang2025entropy}. Existing chunking methods, ranging from heuristic rules based on fixed-length and sentence boundaries \cite{lyu2024crud,gao2023retrieval} to the use of semantic similarity models \citep{xiao2023c,RetrievalTutorials}, and even the recent approaches based on LLMs \citep{duarte2024lumberchunker,zhao2025moc}, generally treat chunking as an isolated and passive pre-processing step, rather than an active and forward-looking deep understanding process \cite{merola2025reconstructing}. This flaw is particularly prominent when dealing with domain documents that are rich in terminology and have strong context dependencies. It can easily lead to semantic fragmentation, making it difficult to effectively utilize key information within text chunks. Typical manifestations include: (a) Lack of term definitions: A text chunk references professional terms, abbreviations, or symbols, but their definitions or explanations are located outside the chunk \citep{faber2016specialized,august2023paper}. (b) Lack of background knowledge: The prerequisite assumptions, background knowledge, or global settings required for understanding the text chunk are not included \citep{chen2024supplementing}. (c) Disruption of context dependencies: The logical reasoning within the chunk depends on key information from the preceding or following text. When these knowledge fragments are injected into LLMs, they not only fail to effectively enhance the models' performance but also interfere with their internal reasoning processes \citep{sheng2025dynamic}.

To break through the performance bottleneck in current RAG systems caused by poor text chunking, this paper proposes the \textbf{QChunker} framework for domain knowledge base construction. By simulating a pre-learning phase without specific task guidance, it encourages the model to go beyond simple text boundary identification and instead pursue a mastery of the deep logic and potential knowledge associations of the content. To achieve this goal, QChunker models text chunking as a compound task of text segmentation and knowledge completion. Given that questions are the key to unlocking deep insights \cite{gregersen2018questions}, we design a multi-agent debate framework composed of a question outline generator, text segmenter, integrity reviewer, and knowledge completer. This framework simulates the thinking process of experts when reading documents, autonomously segmenting and constructing high-quality text chunks from the original documents.

In addition, addressing the drawbacks of existing chunking evaluation methods, which rely excessively on downstream tasks, have long evaluation chains, and are inefficient, we propose a new direct evaluation metric called \textbf{ChunkScore}. Unlike previous evaluation methods that rely on indirect signals, ChunkScore defines the quality of an ideal chunking scheme from two dimensions. At the micro level, it emphasizes logical independence, requiring clear boundaries between adjacent text chunks so that they form independent semantic units. At the macro level, it focuses on semantic dispersion, demanding that the entire set of text chunks comprehensively covers the core themes of the document in a low-redundancy manner. This metric has been proven through both theoretical and experimental validation to be capable of efficiently and accurately judging the quality of text chunks. More importantly, ChunkScore is not only an independent evaluation tool but also plays a crucial decision-making role in our QChunker framework. During the text segmentation stage, we use the document outline to conduct multi-path sampling to generate multiple candidate chunking schemes and ultimately rely on ChunkScore as the adjudicator to autonomously select the optimal solution, based on which we train small language models (SLMs) to implement the debate process. To further verify the effectiveness of the QChunker framework in real-world proprietary domains and contribute a domain-specific evaluation benchmark to the community, we construct a hazardous chemical safety dataset. 

To summarize, the contributions of our work are highlighted by the following points: (1) We propose QChunker, a text chunking framework based on multi-agent debate. It combines text segmentation and knowledge completion, aiming to address the issue of semantic fragmentation in domain RAG. (2) We design and validate ChunkScore, an efficient direct evaluation metric for the quality of text chunks, which is free from dependence on downstream tasks. (3) We construct and release a QChunker dataset containing 45K high-quality samples, as well as a small-domain QA dataset focused on hazardous chemical safety, providing valuable resources for community research.
(4) Through benchmark tests in four heterogeneous domains, we systematically demonstrate the key value and potential of the QChunker framework in enabling RAG to serve specific domains.

\section{Related Works}
\subsection{Text Segmentation}
Text segmentation is a foundational task in Natural Language Processing that involves partitioning text into meaningful, coherent units. This process is a critical prerequisite for downstream applications such as information retrieval \cite{li2020neural,wu2024generative,kim2025syntriever} and text summarization \cite{lukasik2020text,cho2022toward,ma2022multi}. A prominent approach leverages topic modeling to identify thematic shifts, which serve as segmentation boundaries \cite{kherwa2020topic,barde2017overview}. This category spans from classical probabilistic methods \cite{blei2003latent} to modern embedding-based models like Top2Vec \cite{angelov2020top2vec} and BERTopic \cite{grootendorst2022bertopic} that capture semantic relationships. Alternatively, text segmentation can be framed as a sequence labeling task. \citet{zhang2021sequence}, for example, utilize BERT to generate contextualized sentence representations and predict a segmentation decision after each sentence, thereby modeling long-range dependencies. Other strategies focus on document structure. Heuristic methods, implemented in frameworks like LangChain \cite{langchain}, employ techniques such as character-level, delimiter-based, and recursive splitting. While effective at preserving structural integrity, these approaches often lack deep contextual understanding. To overcome this limitation, semantic segmentation \cite{RetrievalTutorials} utilizes text embeddings to group semantically coherent passages, identifying boundaries at points where embedding distances indicate a significant topic change.

\subsection{Text Chunking in RAG}
Powered by complex internal architectures and reasoning mechanisms, LLMs have demonstrated exceptional performance on a wide array of language tasks \cite{kim2025syntriever,zhao2025moc}. The RAG framework enhances these capabilities, particularly in knowledge-intensive domains, by augmenting the LLM's input with externally retrieved text chunks \cite{asai2024self,guo2024lightrag,edge2024local,yang2025knowing}. The efficacy of RAG is fundamentally dependent on the quality of its text chunking component. Suboptimal segmentation strategies can introduce incomplete context or irrelevant noise, which adversely affects the performance of downstream QA systems \cite{yu2023chain}. While conventional methods operate at the sentence or paragraph level \cite{lyu2024crud,gao2023retrieval}, more sophisticated approaches have been proposed. For instance, \citet{chen2023dense} introduce the Proposition as a retrieval granularity, defining them as atomic units of fact. Although this approach is highly effective for fact-based corpora like Wikipedia, its fine-grained nature can disrupt the contextual cohesion of narrative texts, leading to information loss. Another advanced technique, LumberChunker \cite{duarte2024lumberchunker}, leverages an LLM to iteratively identify segmentation points. While this demonstrates the potential of LLM-driven chunking, its performance relies heavily on the model's instruction-following fidelity and incurs significant computational costs. Following this trend, MoC \citep{zhao2025moc} enables intelligent chunking with smaller models, yet it does not fully address the issue of semantic fragmentation and lacks a deep semantic understanding of the resulting text chunks.

\section{Text Chunking for Domain RAG}

\subsection{Problem Definition}
We define the text chunking task for domain-specific RAG as a composite function $F$, which maps an original domain-specific document $D$ to a set $C^*$ composed of high-quality text chunks. This process can be articulated as the functional composition of two core subtasks, namely text segmentation and knowledge completion, expressed as:
$F = f_{\text{com}} \circ f_{\text{seg}}.$

\textbf{Text Segmentation ($f_{\text{seg}}$)}: This function aims to partition the original document $D$ into an optimal initial set of text chunks $C$. Let $\mathcal{P}(D)$ denote the set of all possible partitionings of document $D$. A specific partitioning $C \in \mathcal{P}(D)$ is a collection of $n$ text chunks $C = \{c_1, c_2, \dots, c_n\}$ such that $\bigcup_{i=1}^{n} c_i = D$ and for any $i \neq j$, $c_i \cap c_j = \emptyset$. The optimal partitioning $C_{\text{opt}}$ is determined by an evaluation function $\Phi$:
$$
C_{\text{opt}} = f_{\text{seg}}(D) = \underset{C \in \mathcal{P}(D)}{\arg\max} \, \Phi(C).
$$
The evaluation function $\Phi(C)$ quantifies the overall quality of the partitioning $C$, with the objective of maximizing intra-chunk semantic cohesion and minimizing inter-chunk coupling.

\textbf{Knowledge Completion ($f_{\text{com}}$)}: This function takes the optimal initial partitioning $C_{\text{opt}}$ as input and enhances each text chunk $c_i$ that may suffer from semantic deficiencies. For each $c_i \in C_{\text{opt}}$, the completion process aims to extract a necessary knowledge set $K_i \subset D \setminus c_i$ from the global context of the original document $D$ and fuse it with the original text chunk to generate a semantically enhanced text chunk $c'_i$. This process can be represented as:
$
c'_i = c_i \oplus K_i.
$
Here, $\oplus$ represents a non-trivial knowledge integration operator, which is not a simple text concatenation but rather a rewriting operation that maintains semantic coherence and stylistic consistency. Ultimately, the knowledge completion function outputs the enhanced set of text chunks $C^* = \{c'_1, c'_2, \dots, c'_n\}$.

\subsection{Question-Aware Multi-Agent Debate}
\begin{figure*}[t]
    \centering
    \includegraphics[width=\textwidth]{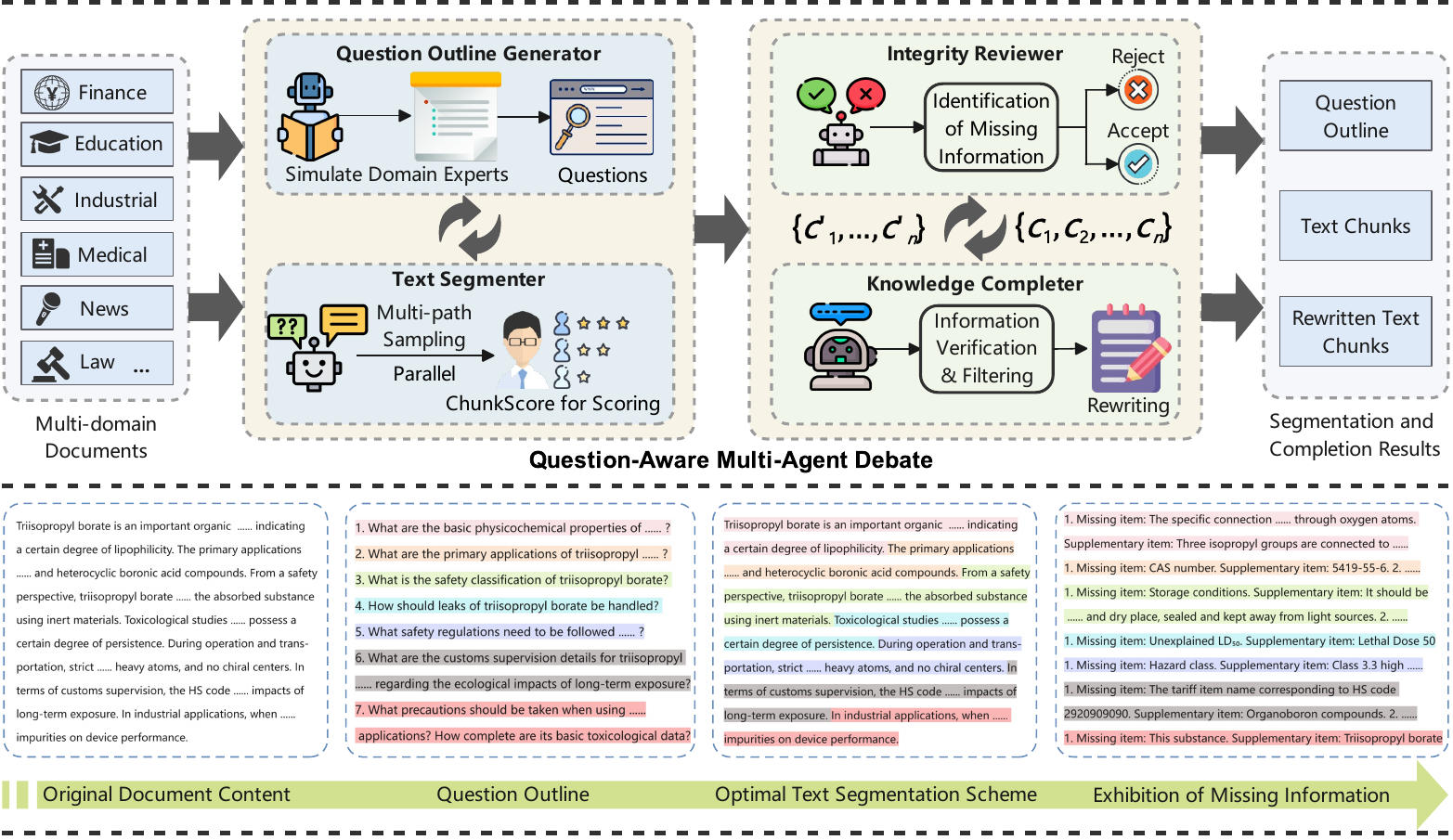}
    \caption{Overview of the QChunker framework's multi-agent debate process. The workflow comprises four key stages: question outline generation, text segmentation, integrity review, and knowledge completion, followed by a concrete example.}
    \label{fig:rag_pipeline}
\end{figure*}

To realize the composite task $F$ defined above, we design a question-aware multi-agent debate framework, the overall architecture of which is illustrated in Figure \ref{fig:rag_pipeline}. This framework decomposes the complex text chunking task onto a set of specialized agents $\mathcal{A} = \{A_{\text{QG}}, A_{\text{SEG}}, A_{\text{IR}}, A_{\text{KC}}\}$ by simulating the cognitive process of an expert team composed of multiple roles. These agents handle the document sequentially within a structured workflow, collectively approximating the optimal solution $C^*$.

\subsubsection{Question Outline Generator}
$A_{\text{QG}}$ serves as the cognitive starting point of the entire framework, with its core responsibility being to simulate the in-depth analysis and exploration process of domain experts when engaging with a document. This process is not a mere information extraction but rather an endeavor to establish a profound understanding of the intrinsic connections between the macro structure and micro arguments of the document. The functionality of this agent can be modeled as a mapping function that transforms the original document $D$ into a structured question outline $\mathcal{Q}$:
$$A_{\text{QG}}: D \mapsto \mathcal{Q}.$$

These questions are designed to probe into the motivation, core assumptions, methodology, key conclusions, and their underlying logical chains within the document. The process of generating $\mathcal{Q}$ essentially compels the model to transition from a passive text processor to an active knowledge explorer, constructing an abstract understanding of the document's knowledge system through self-inquiry and providing a crucial semantic prior for subsequent segmentation.

\subsubsection{Text Segmenter}
The core task of the $A_{\text{SEG}}$ is to implement an optimal partitioning $C_{opt}$ for the document $D$ based on the abstract knowledge structure constructed by $A_{QG}$:
$$A_{\text{SEG}}: (D, \mathcal{Q}) \mapsto C_{\text{opt}}.$$
Searching directly throughout the entire partitioning space $\mathcal{P}(D)$ is computationally infeasible. Therefore, $A_{\text{SEG}}$ adopts a heuristic search and evaluation strategy, utilizing the question outline $\mathcal{Q}$ to effectively prune the search space.

\textbf{Candidate Space Generation}: By leveraging the semantic prior knowledge provided by the question outline $\mathcal{Q}$, the agent generates a manageable yet higher-quality set of candidate partitionings $\mathbb{S} = \{S_1, S_2, \dots, S_p\}$ from the vast universal set $\mathcal{P}(D)$ through a multi-path sampling strategy, where $\mathbb{S} \subset \mathcal{P}(D)$. This step transforms the original global optimization problem into a selection problem on a highly relevant subset.

\textbf{Evaluation and Selection}: To rank each partitioning scheme $S_j$ in the candidate set $\mathbb{S}$, we introduce a specific evaluation function, ChunkScore, the detailed design of which is presented in Section \ref{ChunkScore}. This function is a computable instance of the abstract evaluation function $\Phi$, denoted as $\Phi_{\text{CS}}$. It takes a complete partitioning scheme $S_j$ as input and returns a scalar score to assess its overall quality. Ultimately, the agent determines its output $C_{\text{opt}}$ by solving the following optimization problem:
$$
C_{\text{opt}} = \underset{S \in \mathbb{S}}{\arg\max} \, \Phi_{\text{CS}}(S).
$$
In this manner, $A_{\text{SEG}}$ transforms a theoretically intractable optimization problem into a tractable process of evaluation and selection on a superior candidate space generated under the guidance of $\mathcal{Q}$, thereby efficiently obtaining a text partitioning scheme that approaches the theoretical optimum.

\subsubsection{Integrity Reviewer}
The key task of $A_{\text{IR}}$ is to diagnose and identify potential knowledge incompleteness issues arising from the text segmentation process. In domain-specific texts, even when segmented along semantic boundaries, the resulting text chunks $c_i$ may become difficult to understand or ambiguous when isolated from their global context. By performing a comparative analysis between $c_i$ and $D$, $A_{\text{IR}}$ identifies potential information gaps.

For each identified information gap, $A_{\text{IR}}$ determines whether the absence of such information substantially leads to inaccuracies or comprehension barriers in the text chunk. Ultimately, a structured identification result of missing information within the text chunk is output, along with a definitive judgment on whether supplementary information is indeed required. This process strictly adheres to one principle: all information to be supplemented must be explicitly stated in the original document $D$, prohibiting any form of information extrapolation or creation. The functionality of this agent can be formalized as:
$$A_{\text{IR}}: (c_i, D) \mapsto (\mathcal{M}_i, b_i).$$

The input consists of a text chunk under review $c_i \in C_{\text{opt}}$ and the original document $D$. The output is a tuple $(\mathcal{M}_i, b_i)$. $\mathcal{M}_i$ represents a set of identified knowledge points that are crucial for understanding $c_i$ but are missing within it: $\mathcal{M}_i = \{m_{i,1}, m_{i,2}, \dots\}$. The core constraint is that for any $m \in \mathcal{M}_i$, $m$ must be explicitly stated in $D \setminus c_i$. $b_i \in \{0, 1\}$ is a Boolean judgment. When $b_i = 1$, it indicates that $c_i$ indeed suffers from inaccuracies due to missing information and requires knowledge completion.

\subsubsection{Knowledge Completer}
$A_{\text{KC}}$ serves as the final stage of the framework, responsible for executing the knowledge integration operator $\oplus$ to generate the final high-quality knowledge unit $c'_i$. Its functionality can be expressed as:
$$A_{\text{KC}}: (c_i, \mathcal{M}_i, b_i=1) \mapsto c'_i.$$
This agent is activated if and only if $b_i = 1$. Its workflow consists of two stages:

\textbf{Information Verification and Filtering}: $A_{\text{KC}}$ first conducts a secondary review of $\mathcal{M}_i$ to verify the necessity and relevance of each piece of information: (a) whether the information is indeed critical for understanding $c_i$; (b) whether the information is explicitly stated in the original document $D$; (c) whether supplementing this information would introduce irrelevant new topics not covered in $c_i$. Based on the verification results, a final subset of knowledge to be supplemented, $\mathcal{M}'_i \subseteq \mathcal{M}_i$, is selected.

\textbf{Rewriting and Optimization}: Rather than performing a simple concatenation of the qualified information, the knowledge completer executes a meticulous rewriting operation. It analyzes the internal syntactic structure and stylistic features of the text chunk $c_i$ to identify the most appropriate position for seamlessly integrating the missing background knowledge or term explanations in the most natural manner. Ultimately, it outputs a complete, high-quality text chunk $c'_i$ that has undergone knowledge completion and optimization.

\subsection{Design and Validation of ChunkScore}
\label{ChunkScore}
To effectively guide the optimization process of the text segmentation agent $A_{\text{SEG}}$, we design a novel evaluation metric named \textbf{ChunkScore}, denoted as $\Phi_{\text{CS}}$, whose value lies in serving as a quantitative criterion for data selection, similar to a reward function in reinforcement learning. Unlike traditional evaluation paradigms that rely on downstream tasks, ChunkScore is a composite metric capable of directly and efficiently quantifying the quality of any text partitioning scheme $C$ from two orthogonal dimensions: the micro-level and the macro-level.

The core idea of ChunkScore is that an ideal partitioning scheme must simultaneously satisfy two conditions: (1) Micro-level: The boundaries between adjacent text chunks must be clear, effectively demarcating independent semantic units, which we refer to as logical independence. (2) Macro-level: The entire collection of text chunks should cover the core information of the document in a low-redundancy manner, which we refer to as semantic dispersion.

\subsubsection{Logical Independence (LI)}
This metric aims to quantify the effectiveness of text chunks as independent semantic units. If the boundary between two adjacent text chunks $c_{i-1}$ and $c_i$ is clear, then the content of $c_i$ should not be highly dependent on the context of $c_{i-1}$. We utilize the perplexity of a language model to quantify this dependency. For any internal boundary in the partitioning scheme $C = \{c_1, c_2, \dots, c_K\}$, we define its logical independence as:
$$\text{LI}(c_i, c_{i-1}) = \frac{\text{PPL}(c_i | c_{i-1})}{\text{PPL}(c_i)},$$
where $\text{PPL}(c_i)$ represents the perplexity of the language model on the text chunk $c_i$ itself, measuring the internal linguistic coherence of $c_i$. $\text{PPL}(c_i | c_{i-1})$ denotes the conditional perplexity of the model on $c_i$ given the preceding text chunk $c_{i-1}$ as context.

When the boundary is clear and the two text chunks are semantically independent, the context provided by $c_{i-1}$ offers limited assistance in predicting $c_i$, and the $\text{LI}$ metric approaches 1. Conversely, when the boundary is ambiguous and the two text chunks are strongly semantically related, $c_{i-1}$ renders $c_i$ highly predictable, and the $\text{LI}$ metric approaches 0.

To obtain the overall logical independence score for the entire partitioning scheme $C$, we calculate the arithmetic mean of all $K-1$ internal logical independence values:
$$\Phi_{\text{LI}}(C) = \frac{1}{K-1} \sum_{i=2}^{K} \text{LI}(c_i, c_{i-1}).$$

\subsubsection{Semantic Dispersion (SD)}
Next, we define the metric for measuring macro-level chunking diversity: Semantic Dispersion. This metric aims to reward partitioning schemes that enable text chunks to be semantically distinguishable from each other and reduce information overlap. Given a pre-trained embedding model $f_{\text{embed}}$, the calculation process for $\Phi_{\text{SD}}(C)$ is as follows.

First, the embedding model $f_{\text{embed}}$ is used to map each text chunk $c_i$ in the partitioning scheme $C$ to a $d$-dimensional real-valued vector space $\mathbb{R}^d$. This process generates a compact semantic representation for each text chunk:
$$
\mathbf{z}_i = f_{\text{embed}}(c_i) \in \mathbb{R}^d,~\forall i \in \{1, \dots, K\}.
$$
All the embedding vectors $\mathbf{z}_i$ of the text chunks are stacked column-wise to form an embedding matrix $\mathbf{Z}$, which comprehensively captures the semantic information of the entire partitioning scheme
$
\mathbf{Z} = [\mathbf{z}_1, \mathbf{z}_2, \dots, \mathbf{z}_K] \in \mathbb{R}^{d \times K}.
$

To measure the relationships among the embeddings of the text chunks, we calculate their Gram matrix $\mathbf{\Sigma}$. To eliminate potential biases inherent in the embedding feature dimensions, we first define a feature centering matrix 
$$\mathbf{J}_d = \mathbf{I}_d -\frac{1}{d} \mathbf{1}_d \mathbf{1}_d^\top,$$
where $\mathbf{I}_d$ is the $d$-dimensional identity matrix and $\mathbf{1}_d$ is the $d$-dimensional vector of all ones. The formula for calculating the Gram matrix 
$$
\mathbf{\Sigma} = \mathbf{Z}^\top \mathbf{J}_d \mathbf{Z} \in \mathbb{R}^{K \times K}.
$$
Each element $\mathbf{\Sigma}_{ij}$ in the matrix $\mathbf{\Sigma}$ measures the semantic similarity between the $i$-th and $j$-th text chunks after feature centering.

Finally, we define $\Phi_{\text{SD}}(C)$ as the normalized log-determinant of the regularized Gram matrix:
$$
\Phi_{\text{SD}}(C) = \frac{1}{K} \log \det(\mathbf{\Sigma} + \alpha \mathbf{I}_K).
$$
where $\alpha$ is a small regularization constant (e.g., $\alpha=10^{-3}$) introduced to ensure matrix positive definiteness, and $\mathbf{I}_K$ is the $K$-dimensional identity matrix. 

According to matrix theory, the determinant of a matrix is equal to the product of its eigenvalues. If we let $\lambda_1, \dots, \lambda_K$ be the eigenvalues of the matrix $\mathbf{\Sigma} + \alpha \mathbf{I}_K$, then $\Phi_{\text{SD}}(C)$ can also be equivalently expressed as the arithmetic mean of the logarithms of the eigenvalues:
$$\Phi_{\text{SD}}(C) = \frac{1}{K} \sum_{i=1}^{K} \log(\lambda_i).$$
This form intuitively reveals the essence of $\Phi_{\text{SD}}(C)$, which rewards partitioning schemes that enable the embeddings of text chunks to have large variances in all feature directions.

\subsubsection{ChunkScore}
The final ChunkScore is a weighted linear combination of the two aforementioned sub-metrics, forming a composite evaluation function capable of balancing macro-level and micro-level considerations:
$$\Phi_{\text{CS}}(C) = \lambda \cdot \Phi_{\text{LI}}(C) + (1-\lambda) \cdot \Phi_{\text{SD}}(C),$$
where the hyperparameter $\lambda \in [0, 1]$ is used to adjust the relative importance between logical independence and semantic discreteness. We provide a detailed exploration of its settings in Section \ref{Exploration of ChunkScore}. In our framework, $\Phi_{\text{CS}}$ serves as the ultimate optimization objective for the text segmentation agent $A_{\text{SEG}}$, guiding it to find the most balanced text partitioning scheme that excels in both logical independence and content diversity.

\subsubsection{Effectiveness of $\Phi_{\text{SD}}(C)$}
Although the validity of $\Phi_{\text{LI}}(C)$ is based on language model perplexity and its principle is relatively intuitive, the construction of $\Phi_{\text{SD}}(C)$ relies on more profound mathematical principles. In this paper, we justify its rationality from a geometric perspective.

A high-quality text partitioning scheme $C_{\text{high}}$ should exhibit semantic distinctiveness and complementarity among its members. Conversely, a low-quality partitioning scheme $C_{\text{low}}$ demonstrates semantic redundancy or overlap among its members. We will prove that $\Phi_{\text{SD}}(C)$ can effectively distinguish between these two cases. Before proceeding with the proof, we first introduce a key mathematical lemmas.

\begin{lemma}[Gram Matrix Determinant and Volume \cite{de2008use}]
\label{lemma1}
For a set of $K$ vectors $\mathbf{V}=\{\mathbf{v}_1, \dots, \mathbf{v}_K\}$ where $\mathbf{v}_i \in \mathbb{R}^d$, the corresponding Gram matrix $\mathbf{G}$ is defined as $\mathbf{G}_{ij} = \mathbf{v}_i^\top \mathbf{v}_j$, i.e., $\mathbf{G}=\mathbf{V}^\top \mathbf{V}$. The determinant of this matrix equals the squared volume of the $K$-dimensional parallel polyhedra spanned by these vectors:  
$$\det(\mathbf{G}) = \text{Vol}_K^2(\{\mathbf{v}_1, \dots, \mathbf{v}_K\}).$$
\end{lemma}

\textbf{Perspective of Geometric Analysis}: We aim to demonstrate that maximizing $\Phi_{\text{SD}}(C)$ is equivalent to maximizing the volume spanned by the embedding vectors of text chunks in the semantic space.

First, we revisit the matrix $\mathbf{\Sigma} = \mathbf{Z}^\top \mathbf{J}_d \mathbf{Z}$ in the core computational formula of $\Phi_{\text{SD}}(C)$. Here, the feature centering matrix $\mathbf{J}_d$ is an orthogonal projection matrix, possessing the properties of symmetry $\mathbf{J}_d^\top = \mathbf{J}_d$ and idempotence $\mathbf{J}_d \mathbf{J}_d = \mathbf{J}_d$. Utilizing these two properties, we can reconstruct $\mathbf{\Sigma}$ as follows:
$$\mathbf{\Sigma} = \mathbf{Z}^\top \mathbf{J}_d \mathbf{Z} = \mathbf{Z}^\top (\mathbf{J}_d^\top \mathbf{J}_d) \mathbf{Z} = (\mathbf{J}_d \mathbf{Z})^\top (\mathbf{J}_d \mathbf{Z}).$$
We define a new projected embedding matrix $\mathbf{Z}_{\text{proj}} := \mathbf{J}_d \mathbf{Z} = [\mathbf{J}_d \mathbf{z}_1, \dots, \mathbf{J}_d \mathbf{z}_K]$.
Each column of this matrix is the projected result of the original embedding vector $\mathbf{z}_i$ after removing the feature mean. Therefore, the expression for $\mathbf{\Sigma}$ can be simplified to
$\mathbf{\Sigma} = \mathbf{Z}_{\text{proj}}^\top \mathbf{Z}_{\text{proj}}$.
This form is precisely the standard Gram matrix of the projected vector set $\mathbf{Z}_{\text{proj}}$. According to Lemma \ref{lemma1}, we can establish the following logical chain: For a high-quality partition $C_{\text{high}}$, its members are semantically dissimilar, and the corresponding projected embedding vectors $\{\mathbf{J}_d \mathbf{z}_i\}_{\text{high}}$ tend to be linearly independent or even orthogonal in the space. The parallel polyhedra spanned by them has a relatively large $K$-dimensional volume $\text{Vol}_K$. For a low-quality partition $C_{\text{low}}$, its members exhibit semantic redundancy, and the corresponding projected embedding vectors $\{\mathbf{J}_d \mathbf{z}_i\}_{\text{low}}$ tend to be linearly dependent, resulting in a flattened parallel polyhedra spanned by them, with its $K$-dimensional volume approaching zero.

Thus, $\text{Vol}_K^2(\{\mathbf{J}_d \mathbf{z}_i\}_{\text{high}}) > \text{Vol}_K^2(\{\mathbf{J}_d \mathbf{z}_i\}_{\text{low}})$, which directly leads to $\det(\mathbf{\Sigma}_{\text{high}}) > \det(\mathbf{\Sigma}_{\text{low}})$. Since the logarithmic function is monotonically increasing, we have $\Phi_{\text{SD}}(C_{\text{high}}) > \Phi_{\text{SD}}(C_{\text{low}})$.

In conclusion, $\Phi_{\text{SD}}$ serves as an effective and theoretically well-grounded indicator for measuring the semantic dispersion of text chunks. Furthermore, in Appendix \ref{Perspective of Information Theory}, we provide an information-theoretic proof based on entropy maximization, which further validates $\Phi_{\text{SD}}$ as a principled metric for semantic dispersion.

\renewcommand{\arraystretch}{1.1} 
\setlength{\extrarowheight}{1pt} 
\begin{table*}[t]
\caption{Main experimental results are presented in four domain QA datasets. ROU. and MET. represent ROUGE-L and METEOR, respectively. The best result is in bold, and the second best result is underlined.}
\label{main-performance}
\centering
\resizebox{\textwidth}{!}{%
\begin{tabular}{lccc|ccc|ccc|ccc}
\toprule
\multirow{2}{*}{\textbf{Chunking Methods}} & \multicolumn{3}{c}{\textbf{CRUD}} & \multicolumn{3}{c}{\textbf{OmniEval}}  & \multicolumn{3}{c}{\textbf{MultiFieldQA}} & \multicolumn{3}{c}{\textbf{HChemSafety}}  \\
 &\textbf{BLEU} & \textbf{ROU.} & \textbf{MET.} &\textbf{BLEU} & \textbf{ROU.} & \textbf{MET.} & \textbf{BLEU} & \textbf{ROU.} & \textbf{MET.} & \textbf{BLEU} & \textbf{ROU.} & \textbf{MET.} \\
\midrule
Original&0.5022 & 0.5654 & 0.7324 & 0.1906 & 0.2254 & 0.3904 & 0.1707 & 0.2315 & 0.3650 & 0.2282 &	0.2054 & 0.3152    \\
Llama\_index&0.5312 & 0.5896 & 0.7449 & 0.1969 & 0.2350 & 0.4040 & 0.1732 & 0.2363 & 0.3726 & 0.2477 & 0.2234 & 0.3213 \\
Semantic Chunking&0.5188 & 0.5823 & 0.7434 & 0.1913 & 0.2240 & 0.3821 & 0.1609 & 0.2191 & 0.3468 & 0.2037 & 0.1891 & 0.2964 \\
\addlinespace[2pt] 
\cdashline{1-13} 
LumberChunker&0.5061 & 0.5701 & 0.7399 & 0.1997 & 0.2375 & 0.4085 & 0.1841 & 0.2426 & 0.3809 & 0.2298 & 0.2124 & 0.3402 \\
MoC MetaChunker&0.5456 & \underline{0.6031} & 0.7546 & 0.2042 & 0.2457 & 0.4141 & 0.1707 & 0.2255 & 0.3512 & 0.2525 & 0.2307 & 0.3449 \\
Qwen2.5-14B&0.5329 & 0.5920 & 0.7502 & 0.2048 & 0.2473 & 0.4160 & 0.1883 & 0.2497 & 0.3827 & 0.2550 & 0.2250 &  0.3237\\
Qwen3-14B&0.5382 & 0.5953 & 0.7531 & 0.1907 & 0.2329 & 0.4080 & 0.1800 & 0.2412 & 0.3759 & 0.2419 & 0.2195 &  0.3311\\
\addlinespace[2pt] 
\cdashline{1-13} 
\addlinespace[2pt]
\rowcolor[rgb]{0.94,0.94,0.94} QChunker-3B & \textbf{0.5552} & \textbf{0.6114} & \textbf{0.7640} & \textbf{0.2193} & \textbf{0.2673} & \textbf{0.4348} & \textbf{0.1970} & \textbf{0.2613} & \textbf{0.4010} & \textbf{0.2792} & \textbf{0.2457} & \textbf{0.3654} \\
\rowcolor[rgb]{0.94,0.94,0.94} \qquad w/o $\mathcal{M}_{Ref}$ & \underline{0.5433} & 0.6014 & \underline{0.7601} & \underline{0.2139} & \underline{0.2555} & \underline{0.4198} & \underline{0.1935} & \underline{0.2549} & \underline{0.3889} & \underline{0.2719} & \underline{0.2427} & \underline{0.3607}\\
\bottomrule
\end{tabular}%
}
\end{table*}

\subsection{Training SLMs for the Chunking Task}
To transfer the powerful text segmentation and knowledge completion capabilities within the multi-agent debate framework to more lightweight models, we specifically train three SLMs. Each SLM utilizes 45K training data and is responsible for a core subtask in the chunking process, thereby reproducing the key functions of the complex framework in a modular manner.

The primary task of the first SLM $\mathcal{M}_{Gen}$ is to directly generate a complete question outline $\mathcal{Q}$ and the optimal text segmentation result $C_{opt}$ from the original document $\mathcal{D}$ in a single step. The final training data is available in the form $(\mathcal{D}, (\mathcal{Q}, C_{opt}))$.

The second SLM $\mathcal{M}_{Disc}$ focuses on diagnosing whether there is information missing in text chunks. Trained as a discriminator, its task is to directly determine, based on $\mathcal{D}$ and $c_i$, whether $c_i$ suffers from inaccuracies or semantic ambiguities due to the absence of necessary contextual information. The third SLM $\mathcal{M}_{Ref}$ is responsible for performing the task of $A_{KC}$. When $\mathcal{M}_{Disc}$ determines that $c_i$ has missing information, $\mathcal{M}_{Ref}$ is activated. Its training process encompasses two interrelated objectives: (1) $\mathcal{M}_{Ref}$ needs to accurately identify the specific parts of $c_i$ where semantic ambiguities arise due to missing information and locate the corresponding supplementary information from $\mathcal{D}$. (2) Subsequently, based on this information, $\mathcal{M}_{Ref}$ rewrites the entire text chunk to generate a new, more comprehensive text chunk $c'_i$. The detailed training process is presented in Appendix \ref{Training Configurations}.

\subsection{Dataset Construction for Hazardous Chemical Safety}
We contribute a high-quality, small-domain hazardous chemical safety (\textbf{HChemSafety}) dataset for domain RAG. This not only reveals the significant value of RAG in specialized domains but also further highlights the generalization capability of the QChunker framework. 

The construction of the HChemSafety dataset begins with collecting a comprehensive list that covers the names of common and high-risk chemicals. We then conduct targeted crawling of relevant documents to form a corpus. The documents that have undergone preliminary screening are further quantitatively scored using a text quality assessment model. We only retain the subset of documents with the highest scores for the subsequent refinement stage. Next, we employ the sliding window to assist LLMs in the refinement of the documents, resulting in the final version of HChemSafety documents. Afterwards, we once again leverage the information extraction capability of LLMs to perform named entity recognition and relation extraction on the documents, thereby constructing a HChemSafety knowledge graph. Finally, by retrieving the knowledge graph, we automatically generate a large number of QA pairs that cover different knowledge points and reasoning levels, forming the final retrieval QA dataset that can be used for a further evaluation of the performance of domain RAG systems. For details on dataset construction, please refer to Appendix \ref{HChemSafety Dataset}.

\section{Experiment and Analysis}

\subsection{Datasets and Metrics}
To evaluate text chunking for domain RAG, this study utilizes four benchmarks: CRUD \citep{lyu2024crud} for the news domain, OmniEval \citep{wang2024omnieval} for finance, the multi-domain MultiFieldQA\_zh \citep{bai2023longbench}, and our proprietary hazardous chemical safety dataset, HChemSafety. CRUD is specifically designed for long-answer generation. OmniEval provides a comprehensive assessment of RAG system quality through manually annotated data spanning 5 task types and 16 financial topics. MultiFieldQA\_zh is a component of the LongBench suite. Performance is quantified using BLEU, ROUGE-L, and METEOR metrics, which respectively assess n-gram overlap, the longest common subsequence, and semantic similarity via synonyms and syntactic variations.

\subsection{Baselines}
QChunker is evaluated against six representative text segmentation methods, categorized as rule-based, semantic-based, and LLM-driven. The rule-based approaches include Original Chunking \citep{langchain}, which divides text into fixed-length segments irrespective of sentence boundaries, and the Llama\_index method \citep{langchain}, which respects sentence integrity while approximating a preset token threshold per chunk. The semantic-based method, Similarity Chunking \citep{xiao2023c}, leverages sentence embeddings to partition text by grouping sentences with high semantic relevance. In the LLM-driven category, we compare against established frameworks like LumberChunker \citep{duarte2024lumberchunker}, which pioneered prompting a model to identify topic shifts, and MoC \citep{zhao2025moc}, a parameter-efficient paradigm using a router and meta-chunkers. Furthermore, to assess the impact of the underlying model, we conduct direct comparisons against the general-purpose LLMs, Qwen2.5-14B \citep{team2024qwen2} and Qwen3-14B \citep{yang2025qwen3}.

\subsection{Implementation Details}
In our methodology, the construction of the core training data and the HChemSafety dataset is facilitated by DeepSeek-R1 \citep{guo2025deepseek}. To foster diversity in the generated content, we set the temperature to 0.7 and top\_p to 0.8. For the training of smaller models, we selected Qwen2.5-3B as the base model. During the model evaluation phase, we primarily utilized Qwen2.5-7B \citep{team2024qwen2}. All language models employed in the experiments are the instruction versions, loaded with float16 precision to optimize computational efficiency. For retrieval-based QA, we construct a vector database using Milvus and choose bge-base-zh-v1.5 \citep{xiao2024c} as the embedding model. We configure top\_k=8 to retrieve the most relevant contextual information. Apart from the differences in text chunking strategies, all other components of the RAG pipeline remain strictly identical. Data processing and the construction of all datasets are carried out using the Huawei Ascend AI technology stack, while model training and evaluation are executed on NVIDIA A800 80G GPUs.

\subsection{Main Experiments}
To comprehensively evaluate the effectiveness and generalization capability of our proposed QChunker framework, we conduct extensive experiments on QA datasets across four different domains. For training QChunker-3B, we use data from the first two datasets, while MultiFieldQA and HChemSafety are treated as out-of-domain benchmarks. The experimental results are presented in Table \ref{main-performance}. The QChunker framework consistently achieves optimal performance across all four datasets. The second-best performance is mostly obtained by the ablation variant w/o $M_{\text{ref}}$ (11/12 metrics) and MoC MetaChunker (1/12 metrics), indicating that the proposed components are non-trivial and each contributes to the final effectiveness. We additionally conducted three independent repeated experiments. QChunker outperforms all baseline methods under a t-test with $p<0.05$. Notably, on the HChemSafety dataset, which is constructed for this study and features a high degree of specialization and term density, the advantage of QChunker is particularly prominent. Its scores not only far surpass those of traditional methods but also outperform chunking strategies based on LLMs. This result indicates that the understanding-retrieval-augmentation paradigm of QChunker can effectively address the challenges of strong contextual dependencies and semantic fragmentation in domain-specific documents. 

To investigate the contributions of the key components within the QChunker framework, we design an ablation experiment by removing the knowledge completion module $\mathcal{M}_{Ref}$. This experiment aims to quantitatively validate the proposition proposed in our paper that "text chunking is a composite task of segmentation and knowledge completion". From the comparison of the last two rows in Table \ref{main-performance}, it can be observed that although the performance of $ w/o~\mathcal{M}_{Ref}$ still surpasses that of most benchmark methods, its performance on all metrics across all datasets is lower than that of the complete QChunker framework. For example, on the OmniEval dataset, the removal of the $\mathcal{M}_{Ref}$ module resulted in a decline in the METEOR score from 0.4348 to 0.4198. This demonstrates that merely performing high-quality text segmentation is insufficient, and semantic compensation and contextual reconstruction through $\mathcal{M}_{Ref}$ are crucial for generating truly information-rich text chunks that are easily comprehensible by downstream models.

\begin{figure}[t]
    \centering
    \includegraphics[width=0.45\textwidth]{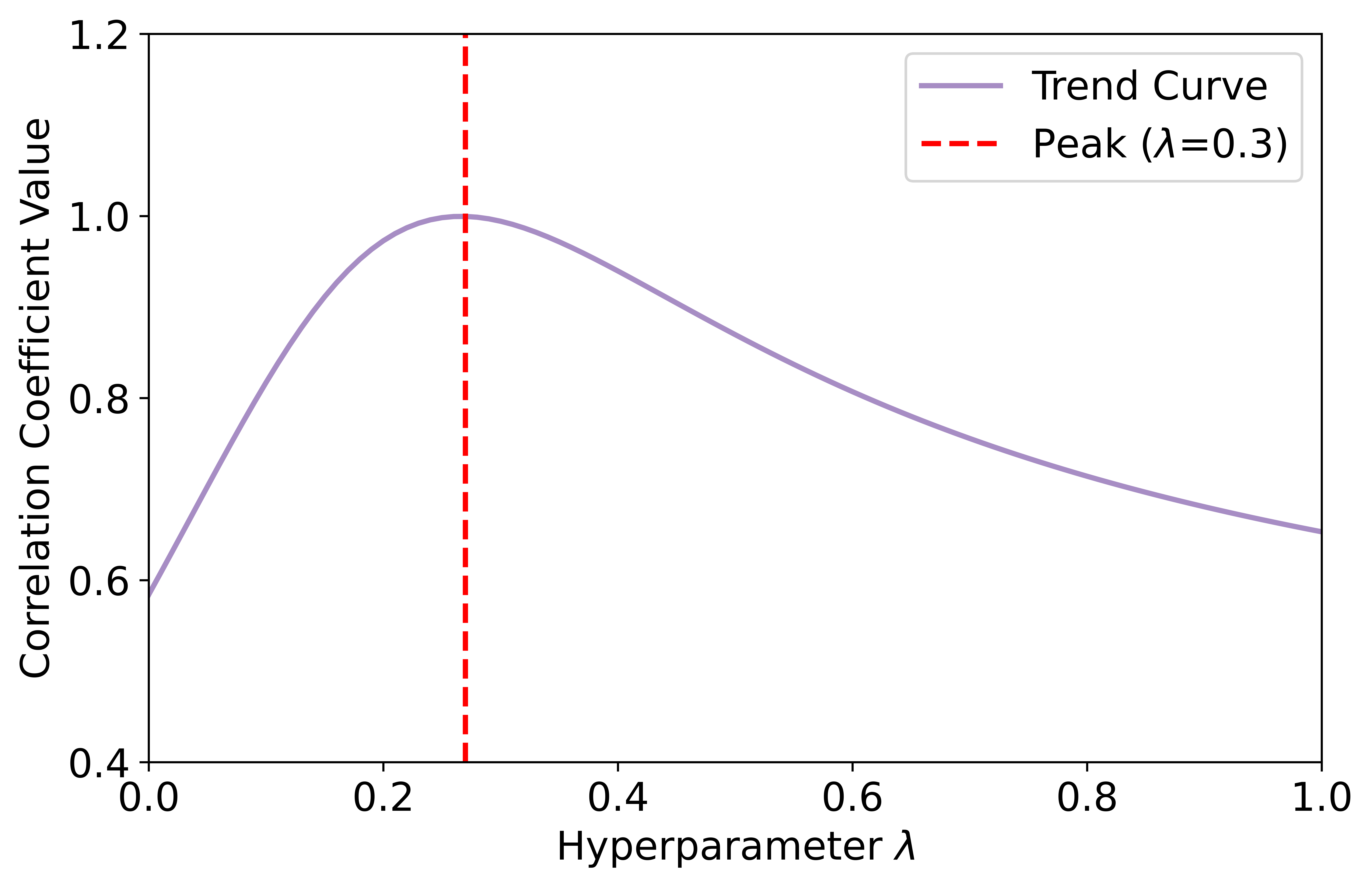}
    \caption{Correlation analysis between ChunkScore and ROUGE-L performance on the CRUD Benchmark.}
    \label{fig:1}
\end{figure}

\begin{figure}[t]
    \centering
    \includegraphics[width=0.47\textwidth]{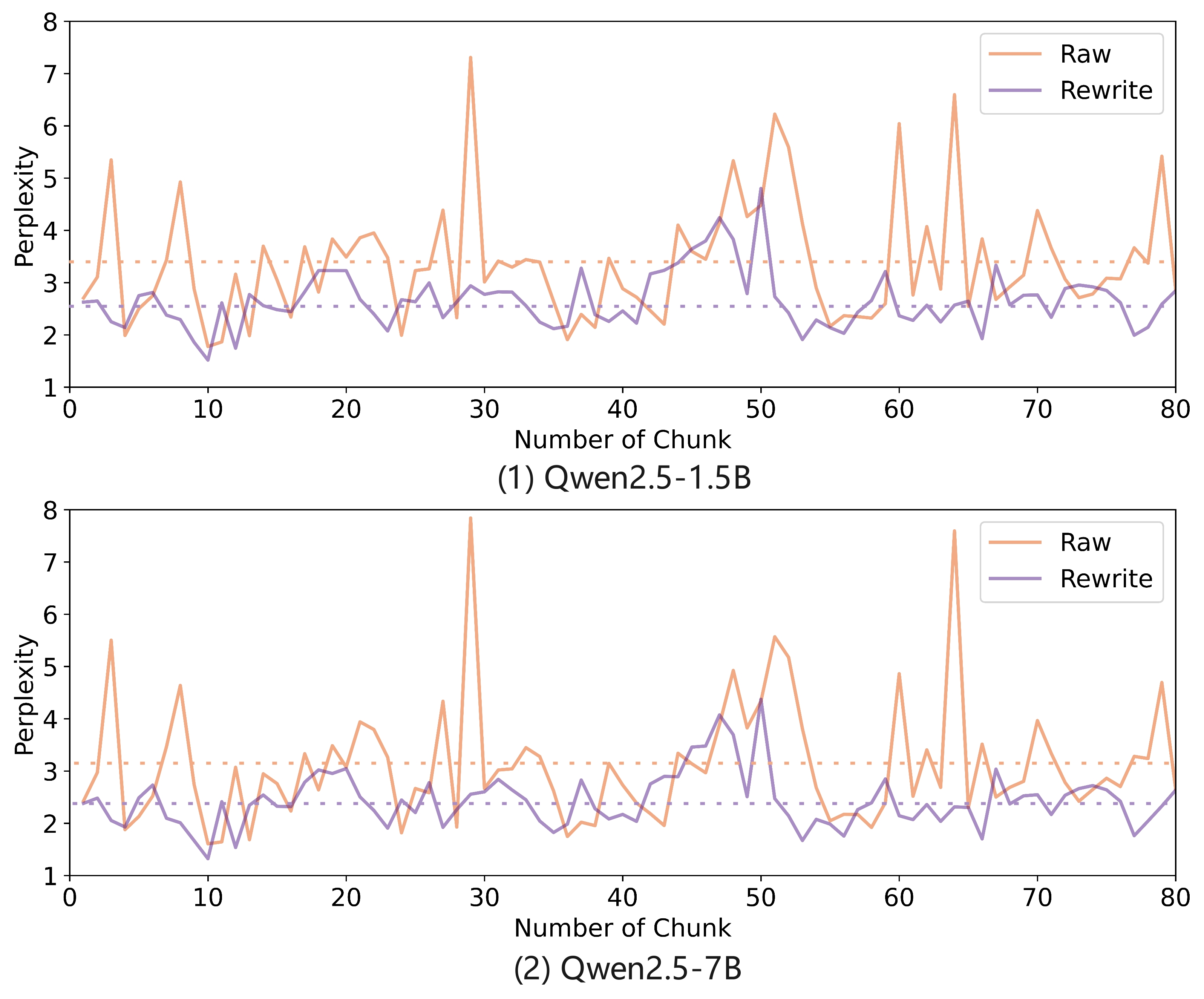}
    \caption{Trends in perplexity variations between original and rewritten text chunks across different LLMs.}
    \label{fig:ppl_rewrite}
\end{figure}

\subsection{Exploration of ChunkScore}
\label{Exploration of ChunkScore}
To further validate the effectiveness of our proposed ChunkScore as a direct evaluation metric and determine the optimal value for its internal hyperparameter $\lambda$, we design a correlation analysis experiment on the CRUD benchmark. We adjust the value of $\lambda$ within the range from 0.0 to 1.0 with a step size of 0.01. For each $\lambda$ value, we calculate the Pearson correlation coefficient between the ChunkScores of multiple text chunk sets generated by different chunking strategies and the ROUGE-L performance in the downstream QA task. The experimental results are shown in Figure \ref{fig:1}. When $\lambda = 0.3$, the correlation coefficient approaches 1.0. This indicates that when ChunkScore considers logical independence with a weight of 0.3 and semantic dispersion with a weight of 0.7, it exhibits the strongest discriminative ability and the highest correlation with QA performance. This also reveals that for high-quality text chunking, ensuring semantic dispersion among text chunks is more important than merely guaranteeing basic boundary clarity, provided that the basic boundary clarity is ensured. We further validate it on three additional datasets (analogous to Figure \ref{fig:1}), and the results show that all correlation coefficients exceed 0.85.

\subsection{Effectiveness of Knowledge Completion}
On the CRUD benchmark, we employ perplexity to more intuitively demonstrate the role of the knowledge completion module in QChunker. Perplexity is a crucial metric for measuring the intrinsic coherence and comprehensibility of text. A lower perplexity indicates that the language model makes more accurate predictions for the text sequence, meaning the logical structure of the text is clearer and more comprehensible for the model.

The experimental results are shown in Figure \ref{fig:ppl_rewrite}. Regardless of whether it is a model with a parameter scale of 1.5B or 7B, the perplexity of text chunks that have undergone knowledge completion is consistently lower than that of original text chunks. Specifically, the text chunks after knowledge completion not only have a lower overall mean value of perplexity but also exhibit less volatility. They avoid many perplexity peaks in original text chunks that are caused by context breaks. This phenomenon proves that the knowledge completion mechanism, by optimizing the internal context consistency of text chunks, pre-eliminates ambiguity points and information gaps that may lead to model comprehension obstacles.

\section{Conclusion}
This paper addresses the core challenge in existing RAG systems due to semantic fragmentation of text chunks, and proposes the text chunking framework named QChunker. Drawing inspiration from the concept that "questions serve as catalysts for deep insights," we design and implement a multi-agent debate framework comprising a question outline generator, text segmenter, integrity reviewer, and knowledge completer. This framework not only constructs a training dataset containing 45K high-quality samples but also transfers this advanced capability to SLMs. Additionally, to address the inefficiency issues caused by the over-reliance of existing evaluation methods on downstream tasks, we introduce a direct evaluation metric called ChunkScore. Both theoretical analysis and experimental results demonstrate that ChunkScore can efficiently distinguish the quality of text chunks. Extensive experimental results across four heterogeneous domains indicate that QChunker outperforms existing methods in providing more logically coherent and information-rich text chunks. In particular, our knowledge completion mechanism effectively reduces the perplexity of text chunks through global semantic compensation. Furthermore, by constructing a specialized dataset in the field of hazardous chemical safety, this study demonstrates the generalization capability of the QChunker framework and its application value in serving specialized domains.

\begin{acks}
This work was funded by the National Natural Science Foundation of China (No.62272466, U24A20233, 72572090), the Beijing Municipal Science and Technology Project (No. 2241100004224009), the Big Data and Responsible Artifcial Intelligence for National Governance of Renmin University of China, the Tsinghua University School of Economics and Management Research Grant. This research was also supported by Huawei’s Al Hundred Schools Program and was carried out using the Huawei Ascend AI technology stack. We also extend our thanks to Renzhe Xu and Hao Zou for their insightful discussions.
\end{acks}

\bibliographystyle{ACM-Reference-Format}
\balance
\bibliography{sample-base}

\appendix

\section{Detailed Model Training Configurations}
\label{Training Configurations}
In this study, we adopt a full-parameter fine-tuning strategy for the three SLMs mentioned previously, namely $\mathcal{M}_{Gen}$, $\mathcal{M}_{Disc}$, and $\mathcal{M}_{Ref}$, with the aim of fully unleashing the potential of the models in their respective specific tasks. To ensure that each model can conduct sufficient learning, we construct training datasets containing 45K samples for them respectively.  We select $1.0 \times 10^{-5}$ as the learning rate and employ a cosine annealing strategy for dynamic adjustment. To stabilize the model's convergence process in the early stage of training, we set a 10\% warm-up phase. Considering the limitations of computational resources, we set the batch size per device to 2 and combine it with gradient accumulation over 16 steps. This approach helps improve the model's generalization ability while ensuring training stability. Additionally, we enable BF16 mixed-precision training to enhance the training speed and reduce the GPU memory usage. The loss function utilized is as follows:
\begin{equation*} 
\mathcal{L}_\text{F}(\theta) = -\frac{1}{\tau} \sum_{t = 1}^{\tau} \log P(o_t | o_{<t}, s; \theta) \label{eq:cross_entropy_loss} 
\end{equation*} 
Here, $o_t$ represents the $t$-th token in the target sequence $o$, $o_{<t}$ denotes the prefix of the target sequence up to position $t-1$, $s$ is the input context, $\theta$ signifies the learnable parameters of the SLM, and $\tau$ represents the total length of the target output sequence $o$.

\section{Main Experimental Details}
In our experiments, we employ a total of five baseline methods, with their specific configurations detailed as follows:
\begin{enumerate}[(a)]
    \item \textbf{Rule-based Chunking Methods}
    \begin{itemize}
        \item \textbf{Original} \citep{langchain}: This method divides long texts into segments of a fixed length, such as two hundred Chinese characters or words, without considering sentence boundaries.

        \item \textbf{Llama\_index} \citep{langchain}: This method considers both sentence completeness and token counts during segmentation. It prioritizes maintaining sentence boundaries while ensuring that the number of tokens in each chunk are close to a preset threshold. We use the \texttt{SimpleNodeParser} function from \texttt{Llama\_index}, adjusting the \texttt{chunk\_size} parameter to control segment length.
    \end{itemize}
    
    \item \textbf{Dynamic Chunking Methods}
    \begin{itemize}
        \item \textbf{Similarity Chunking} \citep{xiao2023c}: Utilizes pre-trained sentence embedding models to calculate the cosine similarity between sentences. By setting a similarity threshold, sentences with lower similarity are selected as segmentation points, ensuring that sentences within each chunk are highly semantically related. This method employs the \texttt{SemanticSplitterNodeParser} from \texttt{Llama\_index}. 
        
        \item \textbf{LumberChunker} \cite{duarte2024lumberchunker}: Leverages the reasoning capabilities of LLMs to predict suitable segmentation points within the text. We utilize Qwen2.5 models with 14B parameters, set to full precision.

        \item \textbf{MoC MetaChunker} \citep{zhao2025moc}: MoC trains a lightweight chunker model to automatically learn how to partition long texts into semantically coherent chunks without relying on fixed lengths or predefined rules. Compared to traditional heuristic methods, MetaChunker demonstrates stronger cross-task generalization capabilities, particularly in downstream tasks such as RAG, serving as the strongest baseline.
    \end{itemize}
\end{enumerate}

To ensure a fair and unbiased evaluation of different text chunking methods, we design a rigorous set of controlled variable experiments. The primary objective of this study is to verify the impact of distinct chunking strategies on downstream task performance in isolation. Therefore, our foremost control strategy is to standardize the output chunk lengths across all methods. Specifically, we adopt the average chunk length of 178 tokens produced by LumberChunker as the basic chunk length.  Given that different chunking algorithms may not precisely align average chunk length to exactly 178 tokens when handling text boundaries, we introduce a secondary constraint on the total context length in retrieval-based QA tasks. This ensures that the cumulative context length fed into the generative model remains strictly fixed at $178 \times 8 = 1424$ tokens. During the final QA generation phase, all methods utilize the same pre-trained language model as the generator.  

By implementing rigorous and uniform control over the aforementioned variables (i.e., the average length of text chunks, the total length of retrieved context, and the generation model), we effectively isolate the text chunking method as the sole experimental variable. This ensures that the final performance comparison can truthfully and objectively reflect the inherent strengths and weaknesses of different chunking methods.

\section{Perspective of Information Theory}
\label{Perspective of Information Theory}
\begin{lemma}[Log-Determinant and Differential Entropy \cite{cai2015law}]
\label{lemma2}
For a random variable $\mathbf{X}$ following a multivariate Gaussian distribution $\mathcal{N}(\boldsymbol{\mu}, \mathbf{\Sigma}')$, its differential entropy $H_{de}(\mathbf{X})$ has the following linear relationship with the log-determinant of the covariance matrix $\mathbf{\Sigma}'$:  
$$H_{de}(\mathbf{X}) = \frac{1}{2} \log\left( (2\pi e)^d \det(\mathbf{\Sigma}') \right) = \frac{d}{2} \log(2\pi e) + \frac{1}{2} \log \det(\mathbf{\Sigma}').$$ 
\end{lemma}

\textbf{Perspective of Information Theory}: We aim to show that maximizing $\Phi_{\text{SD}}(C)$ is equivalent to maximizing the differential entropy of the empirical distribution of the text-chunk embedding set, that is, maximizing its information content and uncertainty.

We regard the set of embedding vectors of text chunks $\mathcal{Z} = \{\mathbf{z}_1, \dots, \mathbf{z}_K\}$ as samples drawn from an underlying multivariate data distribution. In this view, the matrix $\mathbf{\Sigma}$ serves as an estimate of the covariance of this sample distribution.

According to Lemma \ref{lemma2}, there is a positive correlation between $\Phi_{\text{SD}}(C)$ and the differential entropy $H_{de}$. For a high-quality partition $C_{\text{high}}$, its semantic diversity implies that the embedding vectors $\mathcal{Z}_{\text{high}}$ are widely distributed and have high dispersion in the semantic space. Its empirical distribution has high variance and high uncertainty, thus corresponding to a relatively high differential entropy $H_{de}(\mathcal{P}_{\mathcal{Z}_{\text{high}}})$. For a low-quality partition $C_{\text{low}}$, its semantic redundancy implies that the embedding vectors $\mathcal{Z}_{\text{low}}$ are highly clustered in a few regions of the space. Its empirical distribution has low variance and low uncertainty, thus corresponding to a relatively low differential entropy $H_{de}(\mathcal{P}_{\mathcal{Z}_{\text{low}}})$.

Since $\Phi_{\text{SD}}(C) \propto \log \det(\mathbf{\Sigma})$, and $\log \det(\mathbf{\Sigma})$ is a decisive component of the differential entropy, it follows directly that $H_{de}(\mathcal{P}_{\mathcal{Z}_{\text{high}}}) > H_{de}(\mathcal{P}_{\mathcal{Z}_{\text{low}}})$ leads to $\Phi_{\text{SD}}(C_{\text{high}}) > \Phi_{\text{SD}}(C_{\text{low}})$.

\section{An Illustrative Example of the QChunker Framework}
\begin{figure*}[h]
    \centering
    \includegraphics[width=\textwidth]{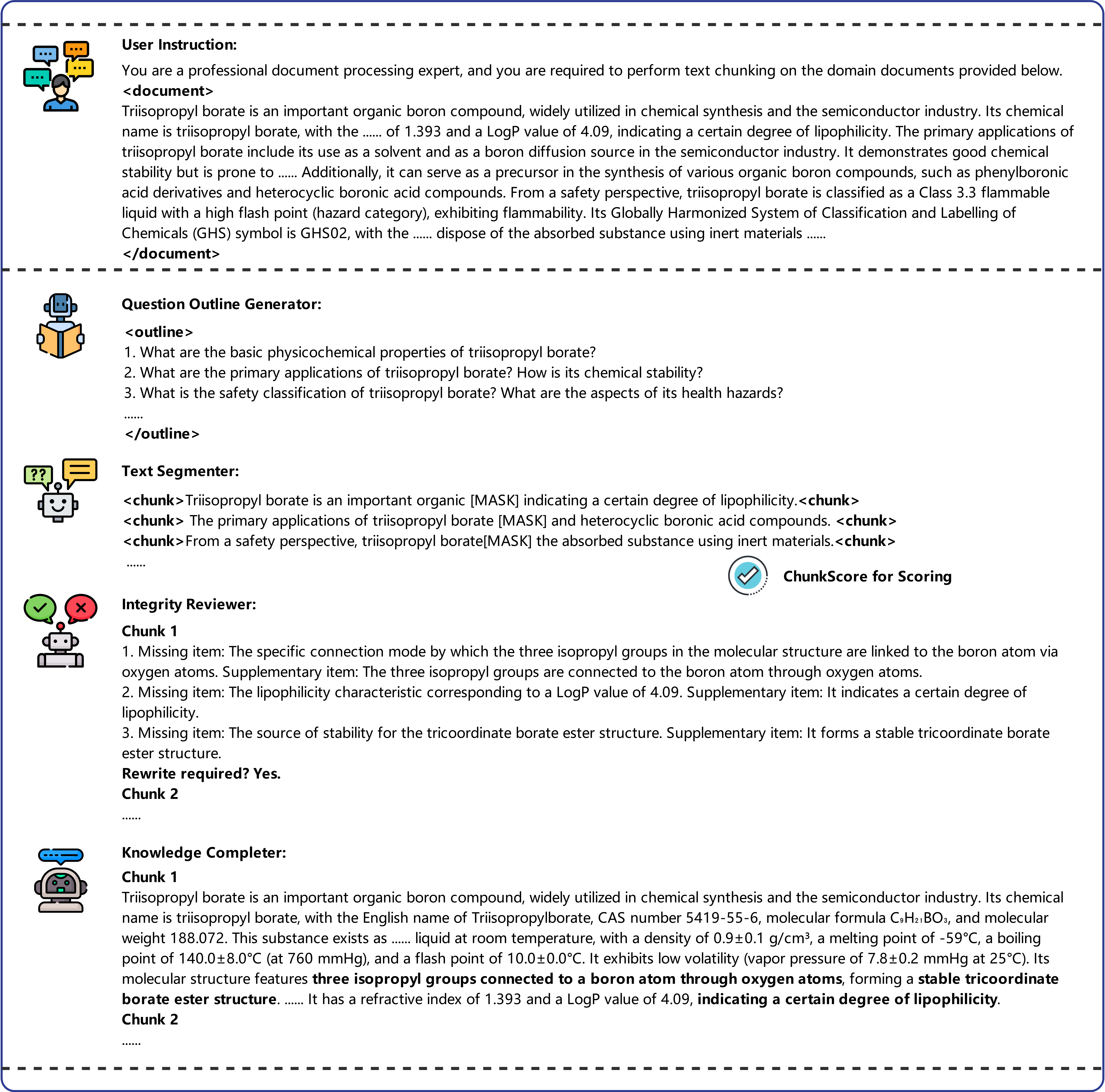}
    \caption{Output example of the QChunker framework when processing chemical documents.}
    \label{fig:rag_pipeline2}
\end{figure*}

Figure \ref{fig:rag_pipeline2} illustrates QChunker’s question-aware multi-agent workflow for building high-quality RAG chunks: given a domain document, a Question Outline Generator first produces expert-style guiding questions that capture the document’s key concepts and logic; a Text Segmenter then uses this outline to perform multi-path sampling over candidate segmentations, scores each candidate with ChunkScore, and selects the best chunking scheme; next, an Integrity Reviewer checks each chunk against the full document to pinpoint missing definitions or required context and decides whether rewriting is necessary; finally, a Knowledge Completer verifies and filters the missing items strictly from the original text and rewrites the chunk to seamlessly integrate essential background, yielding coherent, self-contained, information-rich chunks for downstream retrieval and generation.

\section{In-Depth Construction Process of the HChemSafety Dataset}
\label{HChemSafety Dataset}
To address the challenges of high knowledge barriers and urgent information demands in the field of hazardous chemical safety, we develop the HChemSafety dataset. This dataset aims to fill the gap of a high-quality, domain-specific QA dataset currently lacking in the hazardous chemical safety, providing a solid foundation for evaluating LLM applications.

\subsection{Data Acquisition and Refinement}
The construction of the HChemSafety dataset begins with systematically establishing an initial name repository that comprehensively covers common and high-risk chemicals. Subsequently, by programmatically generating search engine query links, we systematically acquire relevant web page entries, primarily focusing on three types of authoritative platforms: chemical professional platforms, encyclopedic resources, and academic databases. We cluster source URLs based on formatting patterns and design a multi-stage web content cleaning pipeline. First, structural cleaning is performed using DOM tree analysis to identify and remove non-core content areas such as navigation bars, sidebars, footers, and advertising regions, ensuring only the main content area is retained. Second, functional cleaning employs regular expressions and HTML parsers to thoroughly eliminate all interactive elements, including complete removal of <script> and <style> tags, along with stripping all event-handling attributes such as onclick to eliminate dynamic page function interference. Finally, attribute cleaning simplifies the web structure by removing all ID attributes and deleting non-essential class attributes. The cleaned documents then undergo quality assessment. We introduce a fine-grained scoring mechanism based on LLMs \citep{wang2024cci3}, assigning each document a precise score from 0 to 5 and selecting samples with scores greater than 3.

For the filtered high-quality documents, a text refinement process combining sliding windows and LLMs is implemented. This process first segments long documents using text windows with overlapping regions to ensure contextual coherence and prevent critical information loss at boundaries. Each text window is then processed by DeepSeek-V3. Subsequently, the system fuses processing results across all windows, transforming raw web page text into high-value documents.

\subsection{Knowledge Graph Construction and QA Pair Generation}
After completing text refinement, we construct the HChemSafety domain-specific knowledge graph to provide a structured knowledge foundation for subsequent QA pair generation. Knowledge extraction and graph construction constitute the core processes of this workflow. We employ LLMs for entity recognition and relation extraction, achieving end-to-end knowledge acquisition. The extracted triplets are imported into the Neo4j graph database to build the knowledge graph. In this graph, nodes represent entities, while edges denote relationships between entities, thereby more clearly revealing complex interrelations among properties, chemicals, and between chemicals and their properties.  

Based on the constructed knowledge graph, we systematically generate QA pairs covering diverse knowledge points and reasoning levels:  Single-hop QA: Generated directly from a single triplet in the graph. Multi-hop QA: Requires reasoning across multiple edges in the graph. Aggregative QA: Necessitates summarizing information from multiple nodes. Boolean QA: Answers are "yes" or "no". The generated QA pairs are stored in JSON format, including questions, standard answers, and reference content. Each QA pair is directly derived from the knowledge graph, ensuring answer accuracy and traceability.  

Ultimately, we construct a comprehensive hazardous chemical safety dataset comprising three main components: (a) A large-scale QA dataset containing 135K QA pairs, systematically integrated from professional literature, industry standards, and technical manuals to ensure question professionalism and answer authoritativeness. (b) A retrieval corpus consisting of 35K high-quality, clearly structured chemical documents, specifically designed for RAG systems to provide reliable external knowledge sources for fact-checking and answer generation. (c) A curated evaluation set of 19K representative questions, covering various types from basic factual QA to complex safety emergency protocols, enabling assessment of model capabilities in knowledge retrieval, information integration, and specialized reasoning.

\end{document}